\documentclass[conference]{IEEEtran}
\IEEEoverridecommandlockouts

\usepackage{cmap}
\usepackage[T1]{fontenc}
\usepackage[utf8]{inputenc}
\usepackage{newtxtext,newtxmath}
\input{glyphtounicode}
\usepackage{microtype} 
\usepackage[hidelinks]{hyperref}  

\usepackage{cite}                 

\usepackage{amsmath,amssymb,amsfonts}
\usepackage{graphicx}
\usepackage{import}               
\usepackage{xcolor}
\usepackage[caption=false,font=footnotesize]{subfig}
\usepackage{booktabs,multirow,tabularx}
\usepackage{needspace}
\usepackage{tabularx,array}

\usepackage[ruled,vlined,linesnumbered]{algorithm2e}

\usepackage{cmap}
\input{glyphtounicode}
\DontPrintSemicolon
\SetKwInput{KwRequire}{Require}
\SetKwInput{KwEnsure}{Ensure}   
\SetKw{KwDownTo}{down to}
\SetAlFnt{\footnotesize}
\SetAlCapFnt{\footnotesize}
\SetAlCapNameFnt{\footnotesize}
\SetAlgoSkip{0.5em}

\def\BibTeX{{\rm B\kern-.05em{\sc i\kern-.025em b}\kern-.08em
    T\kern-.1667em\lower.7ex\hbox{E}\kern-.125emX}}
    
\begin{document}

\title{Foundation Models for Generalizable Semantic and Goal-Oriented Communication\\
}

\author{
\IEEEauthorblockN{Boliang Liu\IEEEauthorrefmark{1}\IEEEauthorrefmark{2},
Wint Yi Poe\IEEEauthorrefmark{2},
Riccardo Trivisonno\IEEEauthorrefmark{2},
and Giuseppe Caire\IEEEauthorrefmark{1}}
\IEEEauthorblockA{\IEEEauthorrefmark{1}Technical University of Berlin, Berlin, Germany\\
\IEEEauthorrefmark{2}Huawei Technologies D{\"u}sseldorf GmbH, Munich Research Center, Munich, Germany\\
Emails: \{boliang.liu, caire\}@tu-berlin.de;\; \{wint.yi.poe, riccardo.trivisonno\}@huawei.com}
}

\maketitle

\begin{abstract}


Semantic and goal-oriented communication is increasingly studied for 6G, but generalization beyond seen data remains a key weakness under tight rate budgets. Many existing systems overfit their training data and degrade sharply at very low bit rates because they attempt to compress the entire signal. We introduce Foundation Model-Guided Semantic and Goal-Oriented Communication (FMSGOC), a framework that uses broad visual–linguistic Foundation Model priors to mitigate overfitting. It further improves rate efficiency by concentrating bits on sparse, goal-aligned anchors and relying on generative foundation-model priors to reconstruct the masked regions. By decoupling what to send from how to reconstruct, a vision–language foundation model selects and transmits a sparse set of semantic anchors, while a pretrained diffusion model, fine-tuned for masked completion, reconstructs the image at the receiver. In our experiments, FMSGOC reaches 0.039 bits per pixel (BPP), maintains high semantic fidelity (cosine similarity 0.87–0.90 on CIFAR-10), remains robust on previously unseen inputs (0.83–0.86 on ImageNet), and shows good perceptual similarity (0.1278/0.1558, CIFAR-10/ImageNet), outperforming strong end-to-end baselines at lower bit rates.
\end{abstract}

\begin{IEEEkeywords}
Semantic communication, goal-oriented communication, 6G networks, foundation models, diffusion model
\end{IEEEkeywords}

\section{Introduction}
\label{sec:intro}

The transition towards 6G is accompanied by a growing interest in semantic and goal-oriented communication \cite{10634888}, aiming to transmit task-relevant meaning efficiently rather than redundant data. The current dominant paradigm for this task relies on end-to-end (E2E) trained neural systems, such as Deep Joint Source-Channel Coding (DeepJSCC) \cite{8723589} and its variants \cite{Yang2024DiffusionAidedJS,10158995}, which learn to compress an entire input into learned channel symbols for transmission.

However, the dominant E2E paradigm has structural limits: a monolithic model couples semantic feature extraction with compression and reconstruction. As a result, the learned representation is required to encapsulate all information, including fundamental semantics and unimportant data, leading to inefficiency when constrained by strict rate budgets. This limitation results in two issues: (1) the learned features are narrowly tuned to the training data and generalize poorly to unseen content or new objectives \cite{10634888}; (2) at ultra-low bitrates, the information bottleneck becomes too restrictive, and performance degrades sharply because critical semantics cannot be retained. 


These shortcomings reveal that the coupled architecture lacks two key capabilities: (i) the ability to explicitly select task-relevant semantics based on open-world, large-scale knowledge, and (ii) the ability to robustly reconstruct the entire scene from minimal semantic information. Prior work has failed to provide an architecture that overcomes both challenges simultaneously.

In this paper, we propose FMSGOC, a new architecture that breaks from this coupled paradigm. We leverage complementary foundation models (FMs) to independently address semantic selection and generative reconstruction. Our insight is to shift from the lossy compression of the whole scene to the precise transmission of a sparse semantic anchor. This new framework is implemented as follows:
(a) A discriminative FM (a Vision-Language Model (VLM)) performs goal-driven selection to explicitly identify a sparse set of "semantic anchors" based on open-world priors. (b) A generative FM (a Low-Rank Adaptation (LoRA) fine-tuned diffusion model) then leverages its priors for robust reconstruction from these sparse anchors.



This decoupled approach addresses the limitations of E2E systems. By leveraging the FMs' broad pre-trained knowledge, it generalizes zero-shot to unseen data. It also achieves high semantic and perception fidelity at ultra-low bitrates by aiming to preserve core semantics. The results suggest that FM-guided select-and-generate achieves robust performance at ultra-low BPP.

Key Technical Contributions:
\begin{itemize}
  \item A novel semantic communication architecture that decouples goal-driven semantic selection from generative reconstruction, enabling efficient latent-space operations that achieve robust generalization.
  \item A VLM-guided semantic encoder that uses a pretrained VLM to identify a sparse set of latent semantic anchors, enabling goal-oriented rate allocation.
  \item A LoRA fine-tuned diffusion decoder using hard reinjection that performs anchor-preserving masked completion, robustly reconstructing scenes from sparse inputs.
  \item Results on the CIFAR-10 and ImageNet datasets demonstrate that FMSGOC achieves competitive rate-distortion-semantic (R-D-S) performance at ultra-low bitrates, outperforming conventional E2E systems and generalizing to unseen data.
\end{itemize}

\begin{figure*}[!t]
\centering
\includegraphics[width=\textwidth]{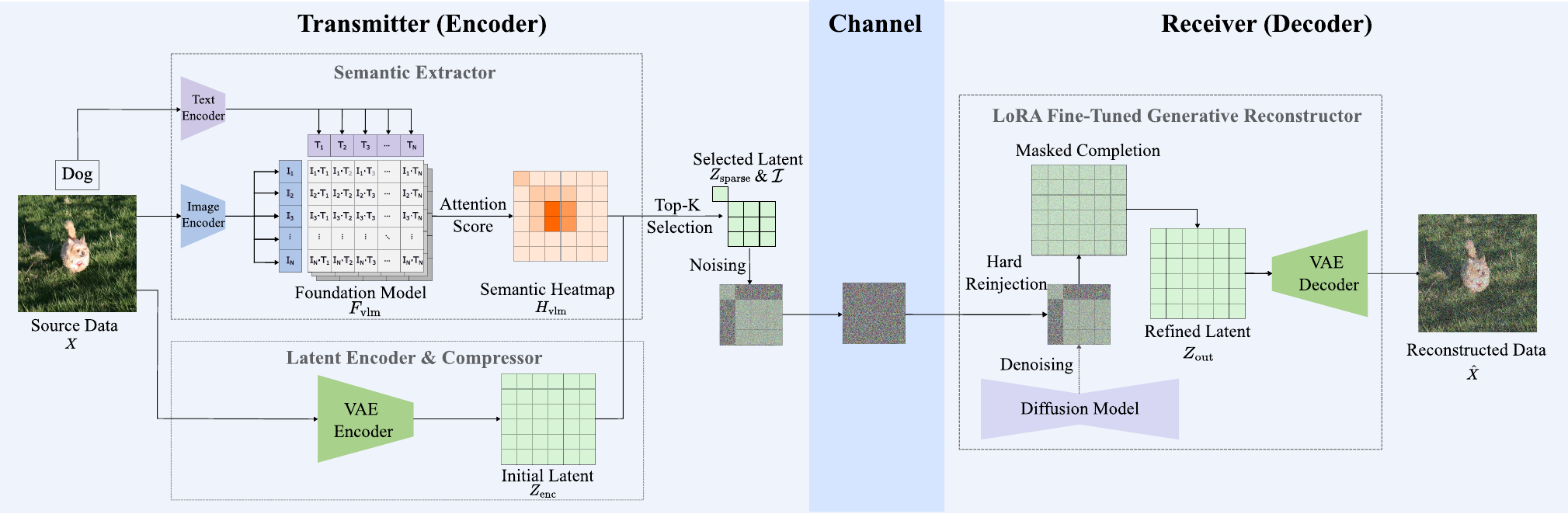}
\caption{The FMSGOC pipeline. \textbf{At Tx,} a VLM-guided Top-K selector extracts sparse semantic anchors ($Z_{sparse}$) and indices ($I$) from a VAE latent grid ($Z_{enc}$). \textbf{Over the channel,} only this ultra-sparse payload ($I, Z_{sparse}$) is transmitted. \textbf{At Rx,} a LoRA fine-tuned diffusion model performs generative masked completion conditioned on the anchors. Hard Reinjection is used during denoising to preserve anchor values, and a VAE decoder renders the final image.}
\label{fig:system_architecture}
\end{figure*}

\section{Related Work}

Recent work on E2E neural codecs (e.g., DeepJSCC\cite{8723589}) jointly optimizes source-channel mappings. These models unify source and channel coding within a pair of encoders and decoders, achieving effective transmission for text \cite{9398576} and image \cite{10094735}. Parallel progress in goal-oriented communication emphasizes transmitting only task-relevant representations to improve the downstream task success rate (e.g., classification or control) rather than pixel fidelity \cite{11129868}; others use a semantic mask for task-aware selective encoding \cite{11202534}. Prior work emphasizes task-relevant features, but most systems remain task-specific E2E models and adapt poorly to unseen data.

Large-scale FMs provide transferable priors across tasks and modalities \cite{Bommasani2021FoundationModels}. This has inspired their use in communication, both as discriminative FMs for semantic guidance and generative FMs for content restoration \cite{10960413}. Others have leveraged generative models like variational autoencoder (VAE) \cite{Kingma2013AutoEncodingVB} or diffusion models to counter severe compression and channel noise at the receiver \cite{Letafati2023ConditionalDD}. While this work validates the capabilities of FMs, they are often used to refine a signal that is already heavily degraded.

\section{System Design}
\label{sec:design}


We decouple the system into three modules in Fig. \ref{fig:system_architecture}: a VLM-guided Semantic Extractor, a VAE Latent Encoder, and a LoRA-adapted generative decoder that performs masked completion from sparse anchors.

\textbf{Transmitter (Tx):}
Given a source image $X$, the transmitter runs two parallel processes:

Latent Encoding (First-Stage Compression): A standard VAE Encoder $E_{\mathrm{VAE}}$ performs the first compression by mapping the image $X$ from the pixel space to a compact initial latent grid $Z_{enc}$.


Goal-Driven Selection (Second-Stage Compression): A VLM functions as a Semantic Extractor, generating a goal-aligned importance map $H_{\text{align}}$ (Section \ref{subsec:encoder}) to select the top-$K$ semantic anchors $(\mathcal{I}, Z_{sparse})$ from $Z_{enc}$.

\textbf{Channel (Payload $\mathcal{C}$):}
The payload $\mathcal{C}$ transmitted over the physical channel consists of two distinct components: (1) the sparse latent anchors $Z_{sparse}$ and (2) their corresponding $K$ grid indices $\mathcal{I}$.

\textbf{Receiver (Rx):}
Upon receiving the payload, the receiver (Rx) initiates a step-wise generative reconstruction process. First, during anchor placement, the received indices $\mathcal{I}$ are used to populate the sparse anchors $Z_{sparse}$ onto an otherwise empty (masked) latent grid. Next, for generative completion, a LoRA-adapted diffusion model $F_{\theta}$ executes its reverse denoising process to fill the unknown (masked) regions. To ensure the output respects the transmitted semantics and prevents "semantic drift," we employ an anchor preservation mechanism that reinjects the received anchor values at each denoising step (detailed in Section \ref{subsec:decoder}). Finally, in the final decoding stage, the completed and refined latent grid $Z_{out}$ is passed to the standard VAE Decoder $D_{\mathrm{VAE}}$ to render the final reconstructed image $\hat{X}$.




\section{Methodology: Foundation Models for Semantic Encoding and Decoding}
\label{sec:methodology}

\subsection{Goal-Driven Semantic Encoder}
\label{subsec:encoder}

The transmitter extracts the minimal set of semantic anchors ($Z_{sparse}$) that satisfy the communication goal. This involves two steps:

\noindent\paragraph{VAE Latent Encoding}
A pretrained VAE encoder $E_{\mathrm{VAE}}$ maps the input image $X$ to a compact latent grid:
\begin{equation}
Z_{\text{enc}} = E_{\mathrm{VAE}}(X), \quad Z_{\text{enc}} \in \mathbb{R}^{H_z \times W_z \times C_z}
\end{equation}


\paragraph{Goal-Driven Anchor Selection}
In parallel, our Semantic Extractor module (Fig. \ref{fig:system_architecture}) uses a frozen VLM (Contrastive Language-Image Pre-Training (CLIP) model \cite{radford2021clip}) to generate a spatial semantic importance map $H_{\mathrm{vlm}}$. This map quantifies the importance of information based on the communication goal. Unlike conventional systems with an implicit goal (i.e., pixel reconstruction), our framework allows this goal to be explicitly defined. 

Unsupervised Mode: For general semantic importance (when no specific goal is provided), $H_{\mathrm{vlm}}$ is generated by aggregating the last-layer self-attention maps from the VLM's vision transformer, as this captures the model's intrinsic focus.


Goal-Driven Mode: When a specific goal is provided, $H_{\mathrm{vlm}}$ is generated by computing the spatial relevance of the goal and the image. We compute the per-patch cosine similarity between the patch representations from the visual encoder and the goal embedding from the text encoder. This leverages CLIP's pre-trained vision-language alignment to estimate the text-conditioned spatial relevance. The subsequent experiments in this paper are based on the goal-driven mode, using the image's category as the prompt.

We then spatially align the importance map $H_{\mathrm{vlm}}$ to the VAE latent grid's resolution ($H_z \times W_z$) via interpolation to produce $H_{\text{align}}$. Finally, we apply a Top-K selection based on the scores in $H_{\text{align}}$, identifying the indices $\mathcal{I}$ of the $K$ highest-scoring cells(we use lowercase $k$ as a percentage, hence $K=\lceil (k/100)\,N\rceil$ with $N=H_zW_z$). The corresponding full latent feature vectors, which span all $C_z$ channels at each selected $(i,j)$ position, are gathered from the VAE grid as the semantic anchors:
\begin{equation}
Z_{\text{sparse}} = \{ Z_{\text{enc}}[i,j,:] \, | \, (i,j)\in\mathcal{I} \}.
\end{equation}
The selection algorithm is detailed in Algorithm \ref{alg:gd_sas}, and the final encoder output pair $(\mathcal{I},\, Z_{\text{sparse}})$.

\begin{algorithm}[!b]
\caption{Goal-Driven Semantic Anchor Selection}
\label{alg:gd_sas}
\KwRequire{Image $X$; pretrained VAE encoder $E_{\mathrm{VAE}}$; frozen VLM $F_{\mathrm{vlm}}$ (e.g., CLIP); selection ratio $k$; optional text prompt $T_{\mathrm{prompt}}$}
\KwEnsure{Index set $\mathcal{I}$ of Top-K anchors; sparse latents $Z_{\mathrm{sparse}}$}

$Z_{\mathrm{enc}}\!\leftarrow\!E_{\mathrm{VAE}}(X)$ \tcp*[r]{latent grid $(H_z,W_z,C_z)$}
$N\!\leftarrow\!H_z\!\cdot\!W_z$\;

\If{$T_{\mathrm{prompt}}$ is null}{
  $H_{\mathrm{vlm}}\!\leftarrow\!\textsc{AggregateSelfAttention}(F_{\mathrm{vlm}},X)$ \tcp*[r]{unsupervised mode}
}\Else{
  $H_{\mathrm{vlm}}\!\leftarrow\!\textsc{CosineSimilarity}(F_{\mathrm{vlm}},X,T_{\mathrm{prompt}})$ \tcp*[r]{goal-driven mode}
}

$H_{\mathrm{align}}\!\leftarrow\!\textsc{BilinearInterpolation}(H_{\mathrm{vlm}},(H_z,W_z))$\;
$K\!\leftarrow\!\lceil (k/100)\,N\rceil$\;
$\mathcal{I}\!\leftarrow\!\textsc{TopKIndices}(H_{\mathrm{align}},K)$ \tcp*[r]{get $(i,j)$ coords}
$Z_{\mathrm{sparse}}\!\leftarrow\!\varnothing$\;

\ForEach{$(i,j)\!\in\!\mathcal{I}$}{
  $Z_{\mathrm{sparse}}\!\leftarrow\!Z_{\mathrm{sparse}}\!\cup\!\{Z_{\mathrm{enc}}[i,j,:]\}$\;
}

\Return $(\mathcal{I},Z_{\mathrm{sparse}})$
\end{algorithm}


\subsection{Generative Decoder}
\label{subsec:decoder}

The receiver's task is to reconstruct the full, coherent image $\hat{X}$ given only the sparse set of constraints $(\mathcal{I},\, Z_{\text{sparse}})$. This is a latent-space image reconstruction task that relies on an Anchor Preservation Triad to address semantic drift.

\label{subsubsec:triad}


\begin{enumerate}
\item \textbf{LoRA Adaptation:} A pretrained generative FM (the diffusion backbone $W_0$) cannot be used directly for this masked-completion task. Standard diffusion models are trained to denoise a full grid from pure noise. When presented with our masked input, they tend to overwrite the known anchor cells, causing severe semantic drift.

Therefore, we must adapt the FM to this new "masked-latent completion" task. We use the lightweight fine-tuning method LoRA for this task. As shown in Fig. \ref{fig:lora_finetuning}, LoRA augments the frozen backbone $W_0$ with small, trainable low-rank matrices ($A, B$):

\begin{equation}\label{eq:lora}
W = W_{0} + \frac{\alpha}{r} B A.
\end{equation}

\begin{figure}[!b]
  \centering
  \includegraphics[width=0.9\columnwidth]{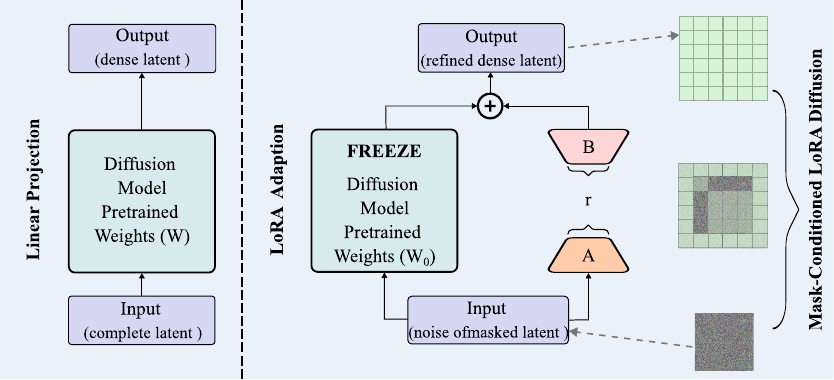}
  \caption{LoRA Adaptation: The frozen diffusion backbone $W_0$ is augmented with lightweight LoRA modules ($A$, $B$).}
  \label{fig:lora_finetuning}
\end{figure}


\item \textbf{Anchor Consistency:}
To prevent semantic drift in known cells, the decoder output is required to be consistent with the transmitted anchors in the masked locations.
Let $M\!\in\!\{0,1\}^{H_z\times W_z}$ denote the anchor mask and $Z_{\mathrm{enc}}$ the input latent grid.
We enforce the constraint
\begin{equation}
\label{eq:anchor_constraint}
M \odot Z_{\mathrm{out}} \;=\; M \odot Z_{\mathrm{sparse}},
\end{equation}
where $Z_{\mathrm{sparse}}$ denotes the received anchor values on the latent grid, and
$\odot$ is the element-wise (Hadamard) product with channel-wise broadcasting, i.e.,
$(M\odot Z)[i,j,:]=M[i,j]\,Z[i,j,:]$.
Unmasked cells $(1{-}M)$ are completed by the fine-tuned diffusion model.

\begin{algorithm}[!b]
\caption{Generative Reconstruction with Hard Reinjection}
\label{alg:recon_hard_reinject}
\KwRequire{Sparse anchor grid \(Z_{\mathrm{sparse}}\); anchor mask \(M\); LoRA-adapted diffusion model \(F_{\theta}\); timesteps \(T\)}
\KwEnsure{Refined latent grid \(Z_{\mathrm{out}}\)}

\tcp*[l]{Initialize \(Z_T\) (noise + anchors)}
\(\epsilon \sim \mathcal{N}(0,\mathcal{I})\)\;
\(Z_T \leftarrow (1-M)\odot \epsilon \;+\; M\odot Z_{\mathrm{sparse}}\)\;

\tcp*[l]{Denoising with hard reinjection}
\For{$t \leftarrow T$ \KwDownTo 1}{
  $\tilde{Z}_{t-1} \leftarrow F_{\theta}(Z_t,\, t,\, M)$ \tcp*[r]{\mbox{propose denoised latent}}
  $Z_{t-1} \leftarrow (1-M)\odot \tilde{Z}_{t-1} \;+\; M\odot Z_{\mathrm{sparse}}$ \tcp*[r]{\mbox{reinsert anchor}}
}
\(Z_{\mathrm{out}} \leftarrow Z_0\)\;
\Return \(Z_{\mathrm{out}}\)
\end{algorithm}

\item \textbf{Hard Reinjection:} We use a per-step projection during inference (Algorithm \ref{alg:recon_hard_reinject}) to enforce semantic fidelity. After the LoRA-adapted model $F_\theta$ proposes a denoised latent state $\tilde{Z}_{t-1}$, we manually reinjection the received anchors $Z_{\text{sparse}}$:

\begin{equation}
\label{eq:reinjection}
Z_{t-1} = (1-M)\odot \tilde{Z}_{t-1} + M \odot Z_{\text{sparse}}.
\end{equation}

This hard reinjection (Eq. \ref{eq:reinjection}) ensures that even if the model momentarily deviates, the transmitted anchors are perfectly preserved in the final $Z_0$, preventing any semantic drift.

\end{enumerate}

Each component serves a distinct purpose: LoRA provides the capability for masked completion, $\mathcal{L}_{anchor}$ enforces this capability during training, and hard reinjection guarantees it during inference.

The final refined latent grid $Z_{out}$ (which is $Z_0$ from the diffusion process) is passed to the pretrained VAE decoder $D_{\mathrm{VAE}}$ to render the reconstructed image $\hat{X}$:
\begin{equation}
\hat{X}=D_{\mathrm{VAE}}\!\left(Z_{\mathrm{out}}\right)
\end{equation}


\subsection{R-D-S Metrics: Rate, Distortion, and Semantics}
\label{subsec:metrics}
Given FMSGOC's goal-driven selection and generative reconstruction, conventional pixel-wise metrics (e.g., PSNR) are incomplete. Our semantic communication system should instead be judged by its bit budget, the fidelity of the reconstruction, and the preservation of semantic meaning. We therefore report three axes: \emph{Rate} ($R$), \emph{Distortion} ($D$), and \emph{Semantics} ($S$).
Let $X$ be the source image, $\hat{X}$ the reconstruction, and $H_X\times W_X$ the image size.

\subsubsection*{Rate ($R$)}
We first define the Source Bits per Pixel (BPP) ($R_S$) as the total payload bits normalized by image size:
\begin{equation}
R_S \;=\; \frac{R_{\mathrm{idx}} + R_{\mathrm{latent}} + L_{\mathrm{CRC}}}{H_X W_X}.
\label{eq:r_src_bpp}
\end{equation}
We then report BPP, which accounts for Low-Density Parity-Check (LDPC) ($r_c{=}2/3$), unless otherwise noted, BPP in this paper refers to this coded BPP:
\begin{equation}
R \;=\; \mathrm{BPP}(\mathcal{C})
 \;=\; \frac{R_{\mathrm{idx}} + R_{\mathrm{latent}} + L_{\mathrm{CRC}}}{r_c\, H_X W_X}.
\label{eq:r_main_coded_bpp}
\end{equation}

The source payload $\mathcal{C}$ contains anchor indices and selected latents.
On a latent grid $H_z{\times}W_z$ with $N{=}H_zW_z$ cells, the encoder selects $K$ anchors (Top-$K$).
The index cost is
\begin{equation}
R_{\mathrm{idx}} \;=\; K\,\big\lceil \log_2 N \big\rceil.
\label{eq:r_idx_def}
\end{equation}
We adopt independent addressing over the combinatorial lower bound \( \lceil \log_2 \binom{N}{K} \rceil \) to prioritize robustness and simplicity at the \(K\) positions; the overall BPP remains ultra-low because only the \(K\) selected latent cells are transmitted and this conservative index cost \(R_{\mathrm{idx}}\) is non-dominant relative to the latent payload \(R_{\mathrm{latent}}\).

For the $K$ latent vectors (channel dimension $C_z$), we use 8-bit adaptive uniform quantization and a discretized Gaussian likelihood to estimate
\begin{equation}
R_{\mathrm{latent}} \;=\; \sum_{i=1}^{K}\sum_{d=1}^{C_z} - \log_2 p(\hat z_{i,d}).
\label{eq:r_latent_def}
\end{equation}
A 24-bit frame-level cyclic redundancy check (CRC) is included as $L_{\mathrm{CRC}}$.


\subsubsection*{Distortion ($D$)}
We use Learned Perceptual Image Patch Similarity (LPIPS)~\cite{8578166} with the standard pretrained AlexNet backbone.
Per-image distortion is
\begin{equation}
D \;=\; \mathrm{LPIPS}(X,\hat{X})
\;=\;
\sum_{l}\alpha_{l}\,\big\|\hat{\phi}_{l}(X)-\hat{\phi}_{l}(\hat{X})\big\|_{2}^{2},
\label{eq:lpips_def}
\end{equation}
where $\hat{\phi}_{l}(\cdot)$ denotes the feature maps at layer $l$ (channel-wise normalized) and $\alpha_l$ are the default LPIPS weights.

\subsubsection*{Semantics ($S$)}
Let $f_{\mathrm{CLIP}}(\cdot)$ denote the frozen CLIP \emph{image} encoder.
Semantic alignment is the cosine similarity in the CLIP embedding space.
\begin{equation}
S \;=\;
\frac{\big\langle f_{\mathrm{CLIP}}(X),\, f_{\mathrm{CLIP}}(\hat{X}) \big\rangle}
     {\|f_{\mathrm{CLIP}}(X)\|_2\,\|f_{\mathrm{CLIP}}(\hat{X})\|_2},
\label{eq:clip_cosine_def}
\end{equation}

\subsection{Training Objective: Composite Loss for Masked Completion}

We fine-tuned the LoRA modules using a composite loss, $\mathcal{L}_{\text{total}}$. This objective is primarily driven by the R-D-S (Rate-Distortion-Semantic) objective. To enforce the LoRA modules to learn to respect our transmitted constraints in Eq.~\eqref{eq:anchor_constraint}, we penalize deviations on masked cells:
\begin{equation}
\label{eq:loss_anchor}
\mathcal{L}_{\mathrm{anchor}}
= \frac{1}{K}\sum_{i,j} M[i,j]\;\big\|Z_{\mathrm{out}}[i,j,:]-Z_{\mathrm{sparse}}[i,j,:]\big\|_2^2,
\end{equation}

Combining this regularizer with the R–D–S terms yields the final objective:

\textit{Total Loss:}
The final objective (Eq. \ref{eq:loss_total}) combines the three primary R-D-S terms with this anchor-consistency stabilizer:
\begin{equation}
\label{eq:loss_total}
\begin{aligned}
\mathcal{L}_{\text{total}}
&= \lambda_{\mathrm{rate}}\,\mathrm{BPP}(\mathcal{C})
 + \lambda_{\mathrm{LPIPS}}\,\mathrm{LPIPS}(X,\hat{X}) \\[-1pt]
&\hspace{0.5em} {}+ \lambda_{\mathrm{sem}}\bigl[
  1 - \mathrm{CosineSim}\!\big(f_{\mathrm{CLIP}}(X),\, f_{\mathrm{CLIP}}(\hat{X})\big)
\bigr] \\[-1pt]
&\hspace{0.5em} {}+ \lambda_{\mathrm{anchor}}\,\mathcal{L}_{\mathrm{anchor}}.
\end{aligned}
\end{equation}

The loss weights were determined empirically based on validation performance. The main R-D-S trade-off is set by $\lambda_{\mathrm{LPIPS}}=0.6$, $\lambda_{\mathrm{sem}}=0.2$, and $\lambda_{\mathrm{rate}}=0.2$. The auxiliary anchor term $\lambda_{\mathrm{anchor}}=0.1$ is set high enough to enforce constraint-awareness in the LoRA modules without overpowering the primary generative optimization.

\section{Performance Evaluation}

\subsection{Experimental Setup}


Experiments are conducted on the CIFAR-10 dataset for training and are evaluated on both CIFAR-10 and ImageNet to assess generalization. 
All experiments (training and evaluation) were performed on H100 GPU with bfloat16. Note that the computing latency is mainly determined by the computing power of the deployed hardware.

The core of FMSGOC relies on two Foundation Models:

\textbf{Goal-Driven Selector (CLIP \cite{radford2021clip}):} Semantic guidance is derived from the pretrained CLIP ViT-B/32 model.

\textbf{Generative Backbone (Stable Cascade \cite{pernias2024wrstchen}):} Our generative pipeline uses the two stages of Stable Cascade. We use the frozen Stage-A VAE ($E_{\mathrm{VAE}}$ and $D_{\mathrm{VAE}}$). Stage-B U-Net as the diffusion backbone ($F_\theta$), adapting it for masked completion using LoRA (see Table \ref{tab:lora_config}).


\begin{table}[b]
\centering
\caption{LoRA Configuration for Stage-B U-Net}
\label{tab:lora_config}
\footnotesize
\begin{tabularx}{\linewidth}{@{}l c >{\raggedright\arraybackslash}X@{}}
\toprule
\textbf{Parameter} & \textbf{Value} & \textbf{Notes} \\
\midrule
LoRA Rank ($r$)        & 32  & Capacity for semantic adaptation \\
LoRA Alpha ($\alpha$)  & 32  & $\alpha/r = 1.0$ scaling \\
LoRA Dropout           & 0.01 & Mild regularization \\
LoRA Clamp             & 0.25 & Stabilizes update magnitude \\
Target Layers          & Conv2d, Linear & Main U-Net compute paths \\
\bottomrule
\end{tabularx}
\end{table}


\subsection{Baselines}
We compare FMSGOC with a classical image compression algorithm and two representative end-to-end (E2E) neural systems. JPEG is implemented using the Pillow encoder as a classical separation baseline. DeepJSCC~\cite{8723589} and WITT~\cite{10094735} use the authors’ official pretrained models on CIFAR-10; all baselines use public checkpoints and author-reported configurations without additional fine-tuning. All methods are evaluated at signal-to-noise ratio $\mathrm{(SNR)}=10$\,dB in an additive white Gaussian noise (AWGN) channel. For separation methods (FMSGOC, JPEG) we report coded BPP under the same LDPC accounting with LDPC rate $r_c=2/3$, while DeepJSCC and WITT are placed on the same coded-BPP axis by matching channel uses. Operating points are produced by using each method’s native rate-control mechanism: FMSGOC varies the retained fraction $k$ of top-$K$ anchors; JPEG varies the discrete cosine transform (DCT) quantization tables; DeepJSCC varies the bottleneck channels (4, 8, 16); and WITT varies the latent channels (8, 16, 24, 32, 48). Markers in the figures denote test-set means at coded BPP.

\subsection{R-D-S Performance and Generalization}

This section evaluates FMSGOC against the baselines on both the in-distribution training dataset (CIFAR-10) and an out-of-distribution dataset (ImageNet) to assess generalization.


\subsubsection{Rate-Semantic (R-S) Performance and Generalization}
\begin{figure}[!t]
\centering
\includegraphics[width=0.9\columnwidth]{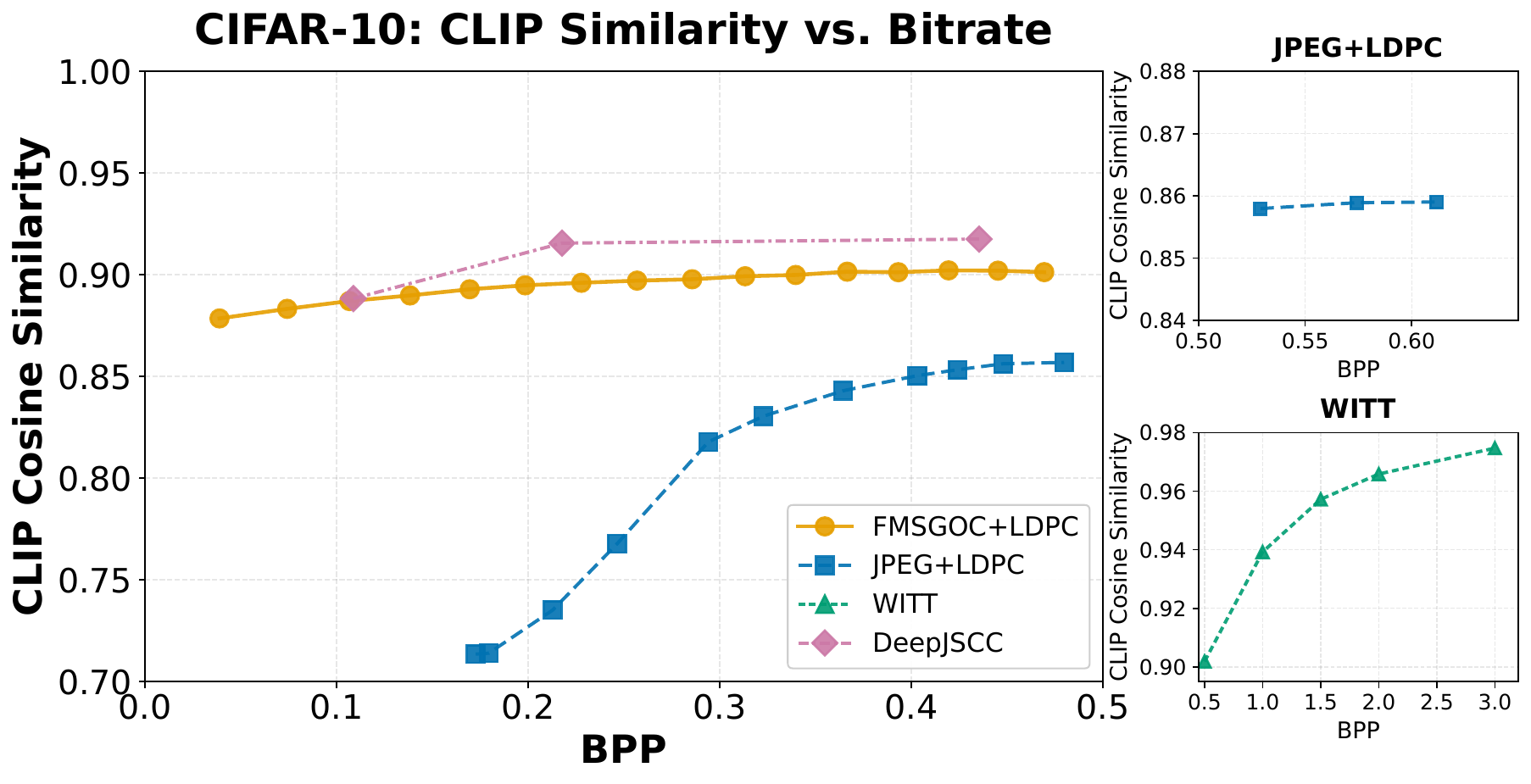}
\caption{CLIP cosine similarity vs. BPP on CIFAR-10.}
\label{fig:cifar10_clip_bpp}
\end{figure}

\begin{figure}[!t]
\centering
\includegraphics[width=0.9\columnwidth]{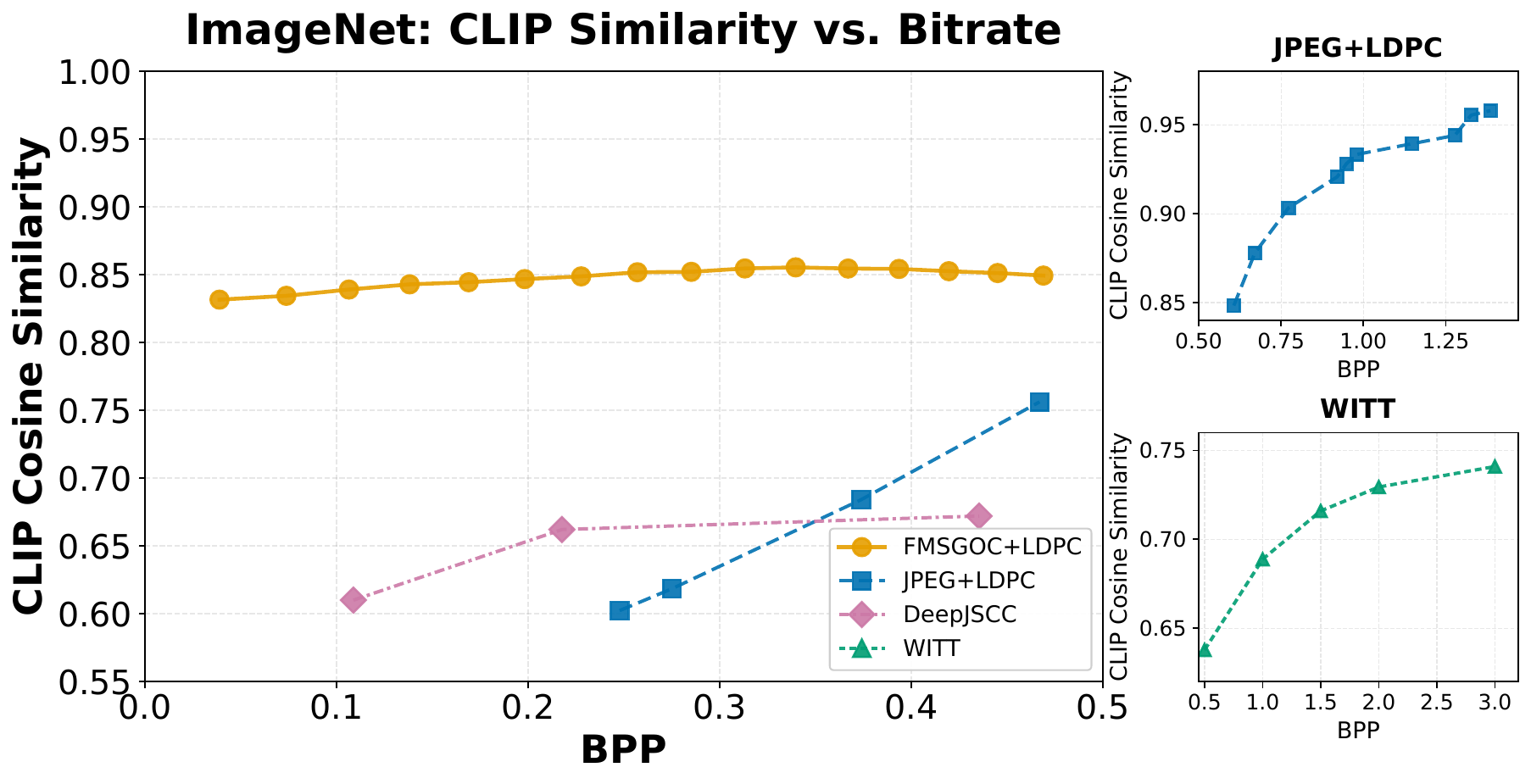}
\caption{CLIP cosine similarity vs. BPP on ImageNet, illustrating semantic generalization under domain shift.}
\label{fig:imagenet_clip_bpp}
\end{figure}

We first analyze the rate-semantic trade-off using CLIP Cosine Similarity (higher is better)(Fig.~\ref{fig:cifar10_clip_bpp} \& Fig.~\ref{fig:imagenet_clip_bpp}).


\textbf{FMSGOC maintains high, stable semantic fidelity and generalizes well.} On CIFAR-10 (in-distribution), FMSGOC's similarity (orange line) remains stable ($\approx$0.87-0.90) even at 0.039 BPP. This preservation of semantics, despite extreme sparsity, is achieved because the VLM selects the most semantically informative anchors, and the generative decoder leverages its pre-trained priors to reconstruct the scene consistent with these anchors. On ImageNet (out-of-distribution), it maintains high similarity ($\approx$0.83-0.86) across all tested rates, confirming zero-shot generalization attributable to the prior knowledge of its FM components.


\textbf{Baselines show poor generalization and are not shown at low rates on CIFAR-10.} On CIFAR-10, the E2E baselines (DeepJSCC, WITT) are plotted only at higher bitrates, where they achieve high similarity. However, their poor generalization is evident on ImageNet. Having been trained only on CIFAR-10, their results show low semantic similarity ($\approx$0.60-0.75) when faced with out-of-distribution data. JPEG is non-semantic and exhibits a sharp degradation in semantic similarity as bitrate decreases on both CIFAR-10 and ImageNet.

\subsubsection{Rate-Distortion (R-D) Performance and Efficiency}
We next analyze perceptual quality (LPIPS, lower is better) vs. BPP (Table~\ref{tab:lpips_bpp_grouped_min}).


\textbf{FMSGOC achieves efficient perceptual quality and generalization.} FMSGOC achieves its minimum LPIPS (0.1278 on CIFAR-10, 0.1558 on ImageNet) at an ultra-low rate of 0.039 BPP.

\textbf{Baselines trade efficiency for quality and fail to generalize.} On the in-distribution CIFAR-10 dataset, all baselines achieve a lower (better) minimum LPIPS than FMSGOC. However, this comes at a significantly higher cost. Crucially, the E2E baselines' (WITT, DeepJSCC) perceptual quality collapses on the out-of-distribution ImageNet dataset, with LPIPS scores ($\ge$0.6217) confirming their lower generalization performance. By contrast, FMSGOC maintains its perceptual quality (0.1558 LPIPS), demonstrating generalization.


\begin{table}[!t]
\centering
\caption{Minimum LPIPS and corresponding bitrate (BPP) on CIFAR-10 and ImageNet for each method.}
\setlength{\tabcolsep}{5pt}
\begin{tabular}{lcccc}
\toprule
& \multicolumn{2}{c}{\textbf{CIFAR-10}} & \multicolumn{2}{c}{\textbf{ImageNet}} \\
\cmidrule(lr){2-3} \cmidrule(lr){4-5}
\textbf{Method} & \textbf{LPIPS (min)} & \textbf{BPP@min} & \textbf{LPIPS (min)} & \textbf{BPP@min} \\
\midrule
FMSGOC   & 0.1278 & 0.0390 & 0.1558 & 0.0390 \\
JPEG     & 0.0667 & 0.6117 & 0.0413 & 1.3272 \\
WITT     & 0.0435 & 3.0000 & 0.6217 & 3.0000 \\
DeepJSCC & 0.1019 & 0.2177 & 0.7234 & 0.2177 \\
\bottomrule
\end{tabular}
\label{tab:lpips_bpp_grouped_min}
\end{table}

\subsection{Ablation Study}
\begin{figure}[!t]
\centering
\includegraphics[width=0.7\columnwidth]{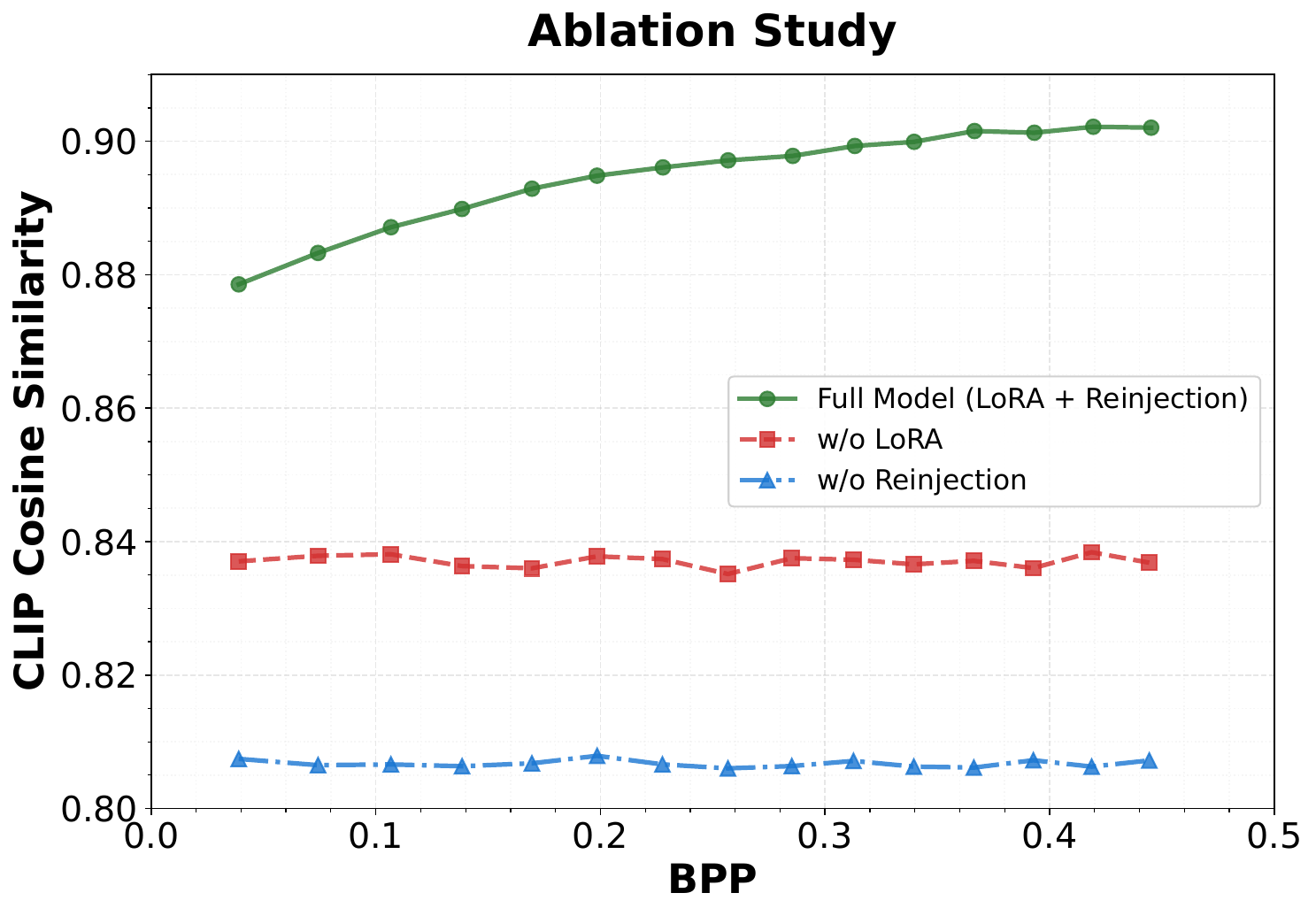}
\caption{Ablation results: CLIP cosine similarity vs. BPP.}
\label{fig:ablation_clip_vs_bpp_low_rate}
\end{figure}

To validate our "Anchor Preservation Triad" (Sec.~\ref{subsubsec:triad}), we present a quantitative, inference-only ablation study (Fig.~\ref{fig:ablation_clip_vs_bpp_low_rate}) comparing our full model against two ablated variants:

FMSGOC (w/o LoRA): The original frozen U-Net without fine-tuning (Eq.~\ref{eq:lora}) is applied. This tests whether task-specific adaptation is necessary.

FMSGOC (w/o Reinjection): The full model, but without Hard Reinjection (Eq.~\ref{eq:reinjection}) during inference. This tests whether the model, once trained, preserves anchors on its own.

\noindent\textbf{Results (Fig.~\ref{fig:ablation_clip_vs_bpp_low_rate}):}
The ablation study confirms that both LoRA adaptation and hard reinjection are essential to the design. The Full Model (CLIP similarity $\approx$0.87-0.90) significantly outperforms both ablated variants, validating that both components are critical, synergistic components:
\begin{itemize}
    \item \textbf{Without LoRA}: Semantic similarity drops significantly (red line, $\approx$0.84). This proves that LoRA adaptation is essential for teaching the model the new masked completion task.
    \item \textbf{Without Reinjection}: Semantic similarity also drops (blue line, $\approx$0.81). This indicates that Hard Reinjection is crucial during inference to prevent the model from "drifting" away from the transmitted anchors.
\end{itemize}

\textbf{Summary of Results:}
The results provide empirical support for our proposed approach: (i) Conventional E2E systems exhibit limitations, showing poor semantic fidelity at low rates on in-distribution data (CIFAR-10) and a sharp degradation in both semantic and perceptual metrics on unseen data (ImageNet). (ii) In contrast, FMSGOC addresses these limitations, demonstrating generalization at an ultra-low 0.039 BPP. (iii) Our ablation study further validates this design, proving that both LoRA fine-tuning and hard reinjection are critical complementary components that respectively enable the masked completion task and prevent semantic drift.


\section{Conclusion}
We have proposed FMSGOC, a decoupled "select-and-generate" architecture built on Foundation Models. This approach targets the key limitations of monolithic E2E systems: poor generalization and performance degradation at low bitrates. Our approach breaks the coupling of feature extraction and reconstruction: a VLM selects sparse semantic anchors, and a LoRA-tuned diffusion model performs generative reconstruction. The results demonstrate robust semantic fidelity and zero-shot generalization at an ultra-low 0.039 BPP, a rate at which E2E baselines show poor semantic fidelity. FMSGOC's ability to address this generalization challenge demonstrates the potential of FM-driven architectures for future 6G networks.

Furthermore, the proposed architecture is inherently controllable and interpretable. Controllability is achieved via explicit Top-K rate selection, while interpretability stems from the goal-driven VLM prompting mechanism. FMSGOC's blend of generalization, controllability, and interpretability is particularly promising for applications such as robotics and autonomous driving, where robust semantic understanding under extreme bandwidth constraints is critical. Future work will explore goal prompt sensitivity and the deployment of distilled, lightweight  models or multimodal models. Another direction is the evaluation of FMSCOC performance on fading channels and channel-aware UEP

\bibliographystyle{IEEEtran}  
\bibliography{references}     %
\end{document}